\documentclass[sigconf,natbib=false]{acmart}
\usepackage{amsmath}
\usepackage{enumitem}
\usepackage{ragged2e}
\usepackage{booktabs}
\usepackage{array}
\usepackage{makecell}

\def\code#1{\texttt{#1}}
\AtBeginDocument{%
  }

\setcopyright{acmlicensed}
\copyrightyear{2024}
\acmYear{2024}

\acmConference[DSAI '24]{Software Development and Technologies for Enhancing Accessibility and Fighting Info-exclusion}{November 13--15,
  2024}{Abu Dhabi, UAE}

\RequirePackage[
  datamodel=acmdatamodel,
  style=acmnumeric,
]{biblatex}
\AtBeginDocument{\renewcommand{\refname}{References}}

\begin{document}

\title{Empowering the Deaf and Hard of Hearing Community: Enhancing Video Captions Using Large Language Models}
\author{Nadeen Fathallah}
\authornotemark[1]
\affiliation{%
  \institution{Analytic Computing, Institute for Artificial Intelligence, University of Stuttgart}
  \city{Stuttgart}
  \country{Germany}}
\email{nadeen.fathallah@ki.uni-stuttgart.de}

\author{Monika Bhole}
\affiliation{%
  \institution{University of Stuttgart}
  \city{Stuttgart}
  \country{Germany}}
\email{st185128@stud.uni-stuttgart.de}

\author{Steffen Staab}
\affiliation{
  \institution{Analytic Computing, Institute for Artificial Intelligence, University of Stuttgart}
  \city{Stuttgart}
  \country{Germany}
}
\affiliation{
  \institution{University of Southampton}
  \city{Southampton}
  \country{UK}
}
\email{steffen.staab@ki.uni-stuttgart.de}

\renewcommand{\shortauthors}{}
\begin{abstract}
In today's digital age, video content is prevalent, serving as a primary source of information, education, and entertainment. However, the Deaf and Hard of Hearing (DHH) community often faces significant challenges in accessing video content due to the inadequacy of automatic speech recognition (ASR) systems in providing accurate and reliable captions. This paper addresses the urgent need to improve video caption quality by leveraging Large Language Models (LLMs). We present a comprehensive study that explores the integration of LLMs to enhance the accuracy and context-awareness of captions generated by ASR systems. Our methodology involves a novel pipeline that corrects ASR-generated captions using advanced LLMs. It explicitly focuses on models like GPT-3.5 and Llama2-13B due to their robust performance in language comprehension and generation tasks. We introduce a dataset representative of real-world challenges the DHH community faces to evaluate our proposed pipeline. Our results indicate that LLM-enhanced captions significantly improve accuracy, as evidenced by a notably lower Word Error Rate (WER) achieved by ChatGPT-3.5 (WER: 9.75\%) compared to the original ASR captions (WER: 23.07\%), ChatGPT-3.5 shows an approximate 57.72\% improvement in WER compared to the original ASR captions.

\end{abstract}



\received{30 June 2024}

\maketitle

\section{Introduction}

The Deaf and Hard of Hearing (DHH) community represents a significant portion of the global population. According to the World Health Organization (WHO), over 5\% of the world’s population – or 430 million people – are estimated to have some degree of hearing loss and require rehabilitation services for disabling hearing loss (including 34 million children). The prevalence of hearing impairment is expected to rise due to factors such as population growth, aging demographics, and increased exposure to harmful noise levels. In particular, it is projected that by 2050, nearly 700 million people will require some form of hearing assistance \cite{39}.

The challenges faced by the DHH community in accessing multimedia content are substantial. Many individuals in this group rely on captions as a primary means of accessing auditory information. However, the quality of captions often falls short of the accuracy required for complete comprehension, significantly affecting the community's ability to fully participate in educational, social, and professional settings. The need to bridge this gap is critical, as video content is increasingly used for education, communication, and entertainment, making accessible captioning an essential aspect of ensuring inclusivity for all.

Although video captions have enhanced accessibility to some extent, there remains a considerable deficit in their precision, user-friendliness, and overall effectiveness. Our research initiative explores the challenges related to the DHH community's use of video captions and improving the quality of captions generated by assistive tools.

Video content is a primary education,  communication, and entertainment medium, facilitating knowledge sharing and communication. Barriers to video content accessibility arise when it has inaccurate captions. Incomplete or inaccurate captions pose a significant challenge for the DHH community, who rely on these captions to access auditory information. Accurate captions are essential to ensure that video content is inclusive and accessible to all individuals, regardless of hearing ability.
Traditional video captioning techniques, such as manual captioning or automatic speech recognition (ASR) systems, often fail to capture the exact spoken words \cite{1}. This inadequacy can lead to distorted meanings or omitting essential information, disrupting content accessibility for the DHH community. 
Our study aims to improve the quality of video captions by integrating Large Language Models (LLMs) into the video captioning pipeline.

\begin{figure}
\begin{center}
    \includegraphics[alt={Example of traditional ASR systems (YouTube's automatic video captioning feature) inaccurate captions. In figure (a) ASR system generates "Koreans canin" instead of "Koreans can", figure (b) "triy" instead of "trial", and figure (c) "business acen" instead of "business acumen".}, width=9cm]{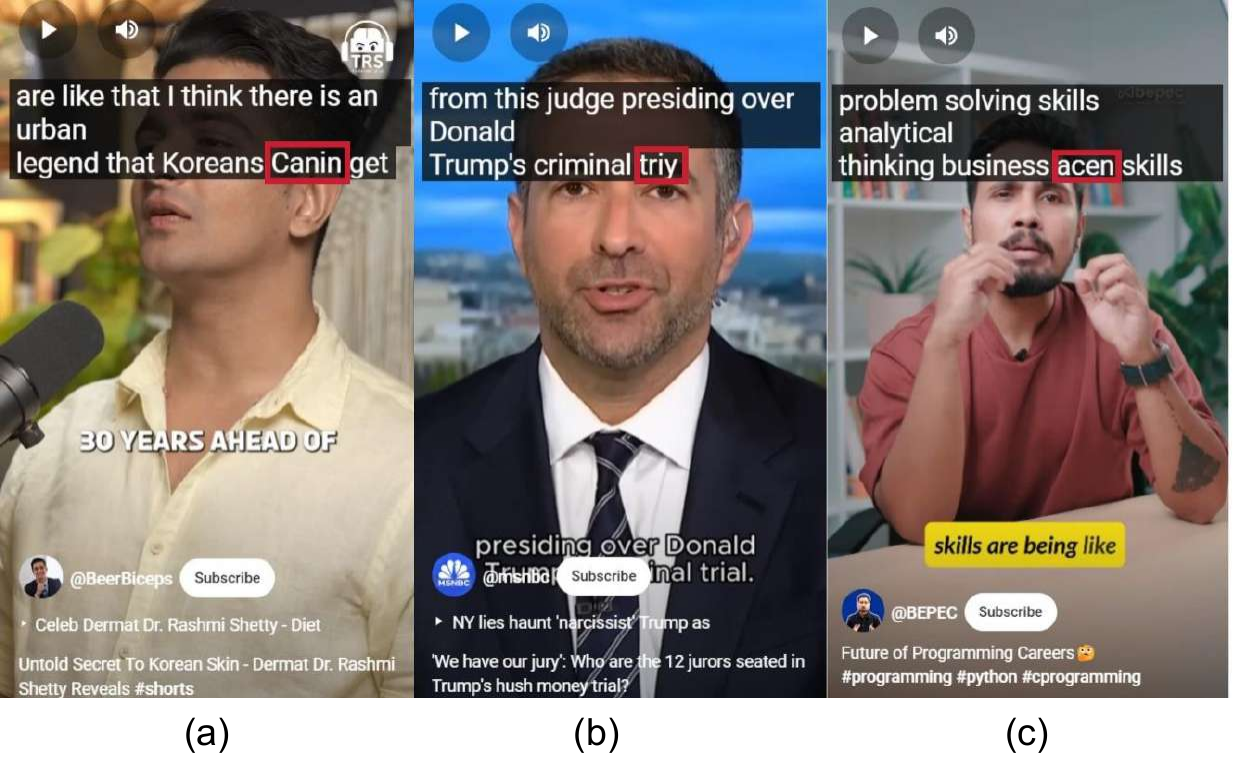}
  \caption{Example of traditional ASR systems (YouTube's automatic video captioning feature) inaccurate captions. In figure (a) ASR system generates \code{"Koreans canin"} instead of \code{"Koreans can"}, figure (b) \code{"triy"} instead of \code{"trial"}, and figure (c) \code{"business acen"} instead of \code{"business acumen"}.}%
    \label{fig:figure1}%
\end{center}
\end{figure}

Figure \ref{fig:figure1} shows inaccurate video captions produced by Youtube's automatic video captioning feature that uses ASR to generate captions. Such errors can alter the content's intended meaning, leading to viewers' confusion. These incorrect captions serve as the input to our proposed pipeline, illustrated in figure \ref{fig:workflow} to improve the quality of video captions, and the corrected captions generated by LLMs are the expected output. The goal of this research is to correct these errors to ensure that video content is fully accessible. 

\noindent Our research question is: 

\begin{enumerate}[label=\textbf{{RQ.}}]
    \item Can Large Language Models (LLMs) be used to enhance the quality of video captions for the DHH community? 

    \end{enumerate}
The primary contributions of this research project are:
\begin{itemize}
    \item Investigating and identifying the issues and challenges the DHH community faces when using video captions.
    \item Using the identified challenges to curate a dataset specifically tailored to these issues.
    \item Leveraging LLMs to improve the quality of ASR-generated video captions produced by existing assistive technologies (e.g., YouTube's automatic video captioning feature).
    \item Evaluating the performance of LLMs in correcting video captions by measuring their accuracy improvements through metrics such as Word Error Rate (WER).
\end{itemize}

\section{Related work}
The DHH community relies on captioning solutions such as ASR systems to access auditory information. This section presents current ASR systems for video captioning, challenges in ASR systems for video captioning, and why LLMs are suitable to address those challenges.

\subsection{ASR Systems for Video Captioning}
Automatic Speech Recognition (ASR) is a technology that uses Machine Learning/Artificial Intelligence to convert human speech into text. It is crucial for the DHH community as it provides a cost-effective, real-time solution for accessing spoken content, enhancing communication, and enabling participation in various activities. Captioning with ASR systems has seen significant advancements in generating high-quality captions, making them a cost-effective alternative to human captioning services \cite{31}, \cite{32}.

Deep learning, particularly end-to-end models like Connectionist Temporal Classification (CTC) and Sequence-to-Sequence (Seq2Seq) architectures, has notably increased transcription accuracy \cite{34}, \cite{33}. Transformer-based models improve performance by handling long-range dependencies in audio sequences \cite{27}. Notable implementations that provide a cost-efficient solution for the DHH community in educational settings include the tabletop tool \cite{37} to facilitate DHH users' communication, the APEINTA system \cite{36} to generate captions on multiple platforms, and the E-Scribe's web-based solution \cite{38}. Despite these advancements, challenges in ASR, such as accuracy in diverse environments, ambient noise, and contextual understanding, remain \cite{35}.

\subsection{Challenges in Video Captioning} \label{challenges}
Despite improvements in ASR technology, the suitability of automatic captions for the DHH community remains a contentious issue. Even minor inaccuracies can significantly impact comprehension, particularly for individuals who rely solely on captions to access auditory information. Many members of the DHH community consider automatic captions to be only a starting point rather than a complete solution. For instance, a seemingly small error rate—such as 0.1\%—could lead to the misinterpretation of critical content, particularly when it involves homophones, specialized terminology, or idiomatic expressions. The reliance on ASR alone without further correction can, therefore, result in incomplete or confusing information. This concern highlights the importance of integrating more sophisticated correction mechanisms to bridge the gap between automatic captions and human-generated quality.

ASR systems still face challenges:

Accuracy Issues: Individuals have distinct accents and dialects, varying from standard American or Indian accents, and regional pronunciation differences. Consequently, video content may contain variations that affect captioning precision, leading to ASR errors \cite{1}, \cite{2}.

Quality of Captions: Captions often suffer from typos, incorrect grammar, and unnatural breaks. ASR systems may misinterpret domain-specific terminology, causing inaccuracies (e.g., confusing "SQL" with "sequel") \cite{1}.

Ambient Noise: Background noise significantly increases captioning errors by impacting speech recognition accuracy. Managing ambient noise is crucial for improving caption precision \cite{1}.

Homophones: ASR systems often confuse homophones (e.g., "to," "two," "too") due to contextual misunderstanding, leading to inaccurate captions \cite{1}.

Speaker Speed: Fast speaking can reduce ASR accuracy, causing captions to miss or misinterpret spoken words \cite{1}.

Code-Switching: Switching between languages or dialects within a conversation challenges ASR systems, requiring seamless handling of multiple linguistic inputs \cite{23}.

\subsection{LLMs for improving Video Captions Quality}
Large Language Models (LLMs) are machine learning models that can comprehend and generate human language text \cite{24}. Earlier studies have typically utilized models like BERT to enhance ASR through methods like rescoring and distillation. Rescoring involves improving the quality of generated text by re-evaluating multiple possible outputs produced by an ASR model. For instance, ASR systems often generate an n-best list—a set of multiple possible transcriptions ranked by likelihood—and a rescoring model can be used to select the most accurate one. For example, \cite{40} applied BERT for n-best list rescoring, while \cite{41} used Minimum Word Error Rate (MWER) loss to train a BERT-based rescoring model that minimizes errors across the top-ranked transcriptions.

Distillation, on the other hand, is a training approach where a smaller, less complex model (the "student") is trained using the knowledge transferred from a larger, more complex model (the "teacher"). This is done by using the teacher's outputs as "soft labels" to provide more informative learning signals. In \cite{42}, BERT was employed in a distillation framework to create such soft labels for training ASR models, effectively teaching the student model nuanced semantic representations. Additionally, \cite{43} explored how to transfer the semantic knowledge embedded in large models, improving ASR performance by enriching the distilled information.

Although these techniques demonstrated the effectiveness of LLMs in improving ASR outputs, they were limited to older LLMs with comparatively less advanced language capabilities.

With the rapid development of newer and significantly more powerful LLMs, such as ChatGPT and Llama2, the potential to enhance ASR systems has increased substantially. These newer models introduce in-context learning, a capability not present in earlier models, which allows them to adapt more flexibly to new prompts without retraining. For example, \cite{44} uses pre-trained LLMs for transforming noisy ASR annotations into high-quality captions, demonstrating the potential of LLMs to generate "human-like" captions without the need for labeled training data. 

Moreover, recent advancements in LLMs make them suitable for addressing the following challenges in video captioning.

Handling Accents and Dialects: LLMs, trained on diverse datasets, adapt to various accents and dialects, enhancing transcription accuracy. \cite{25} highlight that models like GPT-3 can generalize across different linguistic inputs, reducing errors in captions caused by pronunciation variations.

Improving Caption Quality and Accuracy: LLMs enhance grammatical accuracy and coherence. \cite{26} show that GPT-3 generates syntactically correct sentences, which reduces typos and grammatical errors in captions.

Managing Ambient Noise: LLMs leverage contextual understanding to correct words distorted by background noise. \cite{27} introduced the Transformer architecture, enabling models to maintain accuracy even in noisy environments by predicting word sequences based on context.

Differentiating Homophones: LLMs excel at using context to differentiate homophones, reducing errors. \cite{30} demonstrate that models like GPT-3 can distinguish between homophones (e.g., "to," "two," and "too") accurately based on surrounding text.

Coping with Fast Speech: LLMs can handle rapid speech by maintaining contextual coherence, improving transcription accuracy for fast speakers. \cite{28} show that context-aware models like BERT can effectively manage high-speed inputs.

Addressing Code-Switching: LLMs manage code-switching by handling multiple languages within a single context. \cite{29} discuss how models like mT5 enable seamless transitions between languages in captions

\section{Methodology}
\subsection{Dataset}
\subsubsection{Dataset Collection}
\hfill \break While existing datasets like LibriSpeech \cite{14}, TED-LIUM \cite{15}, and Common Voice \cite{16} offer valuable resources for improving ASR system-generated captions, they are not entirely suitable for our task due to their limitations in domain diversity and capturing specific captioning challenges. These datasets often lack the varied contexts and errors in real-world video captions. To address this, we generated our open-domain dataset, comprising videos from diverse domains such as education, cooking, travel and tourism, entertainment, and news. The duration of collected videos ranges from 1-3 minutes. With accessibility often being an "afterthought", we curated our dataset with the DHH community in mind, ensuring it encompasses a wide range of caption challenges users frequently encounter. Our curated dataset includes multiple instances of each of the following common captioning challenges: accuracy and quality issues (e.g., typos, grammatical mistakes, domain-specific terminology), ambient noise, homophones, and speaker speed. 
Our data collection process was driven by the goal of gathering real-world data from a widely used platform that offers an ASR system for generating captions. Recognizing YouTube's extensive use by the DHH community and its ASR system for generating captions as an assistive tool for accessing auditory information, we chose it as our source. Our dataset is publicly available at \url{https://github.com/monikabhole001/Improving-the-Quality-of-Video-Captions-for-the-DHH-Community-Using-LLM}, promoting transparency and enabling further research and development. Details about the dataset are provided in Table \ref{tab:DatasetCollection}. This process resulted in the collection of 52 videos.

\begin{table*}
  \caption{Overview of our dataset collected. The table summarizes the number of videos, the average duration of videos, and the total duration of video content in each domain. "Number of Videos" represents the total count of videos collected in each category. "Average Duration" denotes the mean length of the videos within each category. "Total Duration" indicates the cumulative length of all videos in each domain.}
  \label{tab:DatasetCollection}
  \begin{tabular}{ccccccl}
    \toprule   &Education&Cooking&Travel and Tourism&Entertainment&News&Total\\
    \midrule
   Number of Videos&10&10&10&10&12&52\\
   Average Duration&1m 33s & 1m 57s& 1m 9s & 1m 39s &1m & 1m 27s\\
   Total Duration& 19m 34s& 15m 34s & 11m 31s & 15m 48 s& 11m 28s& 73m 55s \\
    \bottomrule
\end{tabular}
\end{table*}

\subsubsection{Dataset Annotation}
\hfill \break 

Numerous studies have considered ASR an assistive tool for DHH to access auditory information \cite{1}, \cite{2}. However, inaccurate captions are a significant limitation of ASR. After collecting 52 videos from diverse domains, we utilized the \code{YouTubeTranscriptApi} \cite{18} to retrieve the ASR-generated captions from YouTube. Thus, the assistive tool under investigation in this study is YouTube's ASR automatic captioning feature, and the retrieved captions serve as the input to our proposed model. To evaluate our model's caption correction performance, we manually generated the ground truth captions for these videos, enabling us to compare the corrected captions produced by our model to the actual words spoken in the videos. Details about the dataset variables are presented in Table \ref{tab:datasetvariables}

 \begin{table*}
  \caption{Description of our dataset variables. The table includes three columns: "Name" lists the variable names, "Description" provides a brief explanation of each variable, and "Type" indicates the data type of each variable.}
  \label{tab:datasetvariables}
  \begin{tabular}{ccl}
    \toprule   Name&Description&Type\\
    \midrule
  VideoID&A unique identifier assigned to each video in the dataset&integer\\
  URL&The web address where the video can be accessed on YouTube& text\\
  Youtube\_Caption& The captions generated automatically by YouTube's ASR system for the video&text\\
  Ground Truth\_Caption&The manually generated accurate transcription of the video's spoken content&text\\
  Domain& The category or subject area to which the video belongs, such as education, cooking, or news&text\\
  
    \bottomrule
\end{tabular}
\end{table*}

\subsection{Model Selection} \label{model}

This study aims to integrate LLMs with assistive tools commonly used by DHH students to generate and correct automatic captions. The captions produced by these assistive tools serve as inputs for the LLMs, after that we prompt the LLM to correct any inaccuracies. This section details the various LLMs we experimented with to enhance video captioning quality:

\begin{itemize}
    \item GPT-2: OpenAI’s second-generation generative pre-trained transformer model is known for its performance in language modeling and reading comprehension tasks. However, it has shown limitations in summarization and question-answering tasks \cite{3}. Our implementation of GPT -2 for caption correction revealed no significant improvement in the quality of the captions, and it generated irrelevant context for the input caption.

    \item T5: As an encoder-decoder model, T5 excels in translation and summarization tasks. However, it performed poorly in the caption correction task, struggling with understanding context and preserving errors from the input.  \cite{8}.

     \item Llama2-13B: Meta’s LLM, Llama-2, specifically the 7B model \cite{9}. This model showed better handling contextual information and correct captions successfully.
     
    \item GPT-3.5: Representing a significant leap from GPT-2, GPT-3.5 demonstrated superior ability in generating coherent and contextually accurate text \cite{26}. Its advanced processing capabilities resulted in more fluent and precise captions, making it highly effective for enhancing ASR-generated captions. We employed the ChatGPT-3.5 web interface for our corrections.
\end{itemize}

\begin{figure}
\begin{center}
    \includegraphics[alt={Selection process of Large Language Models (LLMs) for improving the quality of captions generated by ASR system. The figure illustrates the performance of different LLMs (GPT-2, T5, Llama2-13B, GPT-3.5) in correcting a sample video caption while maintaining the original word sequence.},width=8.5cm]{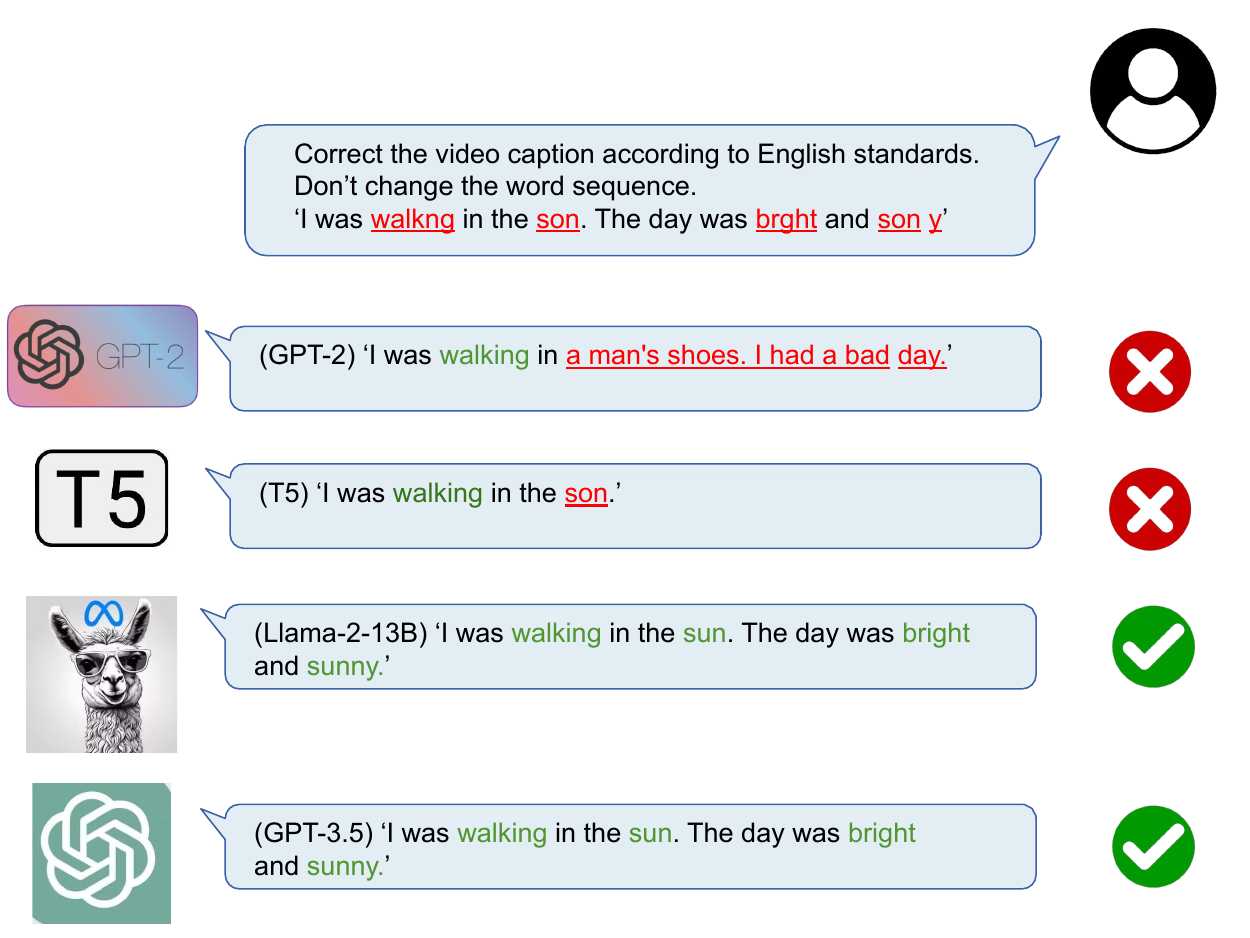}
  \caption{ Selection process of Large Language Models (LLMs) for improving the quality of captions generated by ASR system. The figure illustrates the performance of different LLMs (GPT-2, T5, Llama2-13B, GPT-3.5) in correcting a sample video caption while maintaining the original word sequence.}%
    \label{fig:modelSelection}%
\end{center}
\end{figure}

In our model selection process, we tested various sample captions with errors, one of which is illustrated in figure \ref{fig:modelSelection} \code{'I was walkng in the son. The day was brght and son y.'}. GPT-2 generated irrelevant content, which can be primarily attributed to its nature as a text-generation model. Meanwhile, T5 failed to correct all the caption errors and generate the whole caption. Both Llama2-13B and ChatGPT-3.5 successfully corrected the caption, demonstrating their effectiveness. This comparison highlights the necessity of selecting robust LLMs to improve caption quality in assistive technologies. The notable performance of Llama2-13B and ChatGPT-3.5 led us to incorporate them into our pipeline for further comparative studies to determine the most suitable model for our task.

\subsection{Pipeline for improving video caption quality using LLMs}

To improve the quality of ASR system-generated captions, we propose a pipeline that leverages LLMs' advanced capabilities, including contextual understanding, text generation, text correction, and semantic coherence. Our proposed pipeline is illustrated in figure \ref{fig:workflow}. 

\textbf{Input:} Our dataset of video content covers various domains like education, cooking, travel and tourism, entertainment (such as movies), and news, collected from YouTube. Videos range is between 1 and 3 minutes. We chose YouTube's automatic caption generation feature as the assistive tool under investigation to generate captions. The input to the LLM is the captions generated by YouTube's automatic caption generation feature as text. 

\textbf{Improving captions with LLMs:} We prompt LLMs to correct input inaccurate captions without altering the word sequence. To construct the prompts, we use zero-shot prompting, which refers to the model's ability to perform a task without any prior examples or specific training on that task, based solely on the instructions given in the prompt \cite{13}, with their vast pre-existing knowledge, LLMs can generate accurate and contextually appropriate captions without needing additional task-specific information or training. We conduct experiments with two LLMs, ChatGPT-3.5 and Llama2-13B, based on their performance as discussed in \ref{model}.

\textbf{Output:} The output generated by LLM is the corrected captions. The output generated by LLM is then compared with manually created ground truth. The generated captions are more accurate and contextually relevant than the initial captions. 

\textbf{Implementation:} We implemented our project using Python within a Jupyter Notebook environment. 

Link to GitHub project repository: \url{https://github.com/monikabhole001/Improving-the-Quality-of-Video-Captions-for-the-DHH-Community-Using-LLM}

\begin{figure*}
\begin{center}
    \includegraphics[alt={Pipeline - Our proposed caption correction pipeline leveraging LLMs. The input is the ASR system-generated caption (text), shown on the left, which includes errors highlighted in red. The output is the LLM-corrected caption (text), shown on the right, where the corrections are highlighted in green.}, width=16cm]{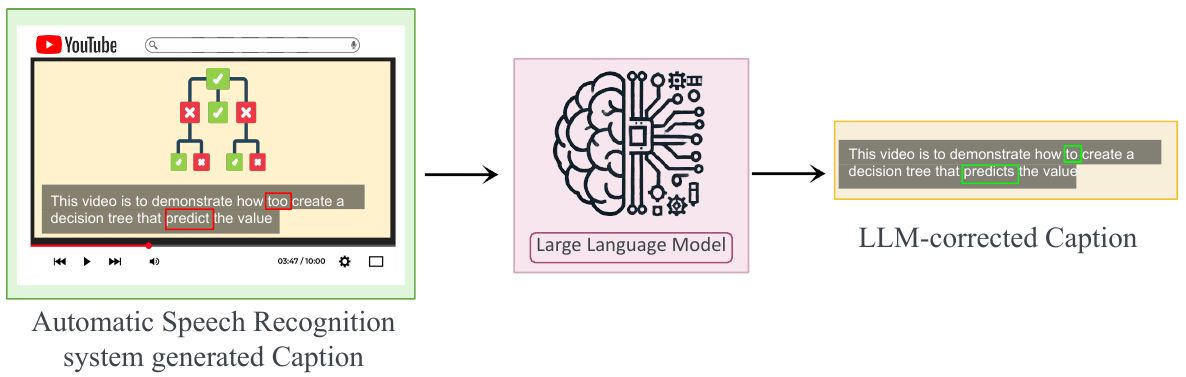}
  \caption{ Pipeline - Our proposed caption correction pipeline leveraging LLMs. The input is the ASR system-generated caption (text), shown on the left, which includes errors highlighted in red. The output is the LLM-corrected caption (text), shown on the right, where the corrections are highlighted in green.}%
    \label{fig:workflow}%
\end{center}
\end{figure*}

\section{Evaluation}

The final step is the evaluation of our caption correction pipeline. We employ two approaches for evaluation: how the selected LLMs (Llama2-13B and ChatGPT-3.5) perform on different challenges in captions, such as homophones, inaccurate words, and domain-specific terminology, and the use of quantitative metrics for a more objective assessment of caption accuracy and quality.

\subsection{LLM Performance on Captioning Challenges} \label{captionChallenge}
Prompt: \code{“Correct the caption according to English standards. Don’t change the word sequence”}.

\begin{itemize}
    \item Homophones and inaccurate words: 
\begin{itemize}
    \item Input caption: \code{"I was walkng in the son. The day was bright and pleasant because of son y weather."}
    \item ChatGPT output caption: \code{"I was walking in the sun. The day was bright and pleasant because of sunny weather."} 
    \item Llama2-13B output caption: \code{"I was walking in the sun. The day was bright and pleasant because of sunny weather."}
\end{itemize} 

\item Domain-specific terminology:
\begin{itemize}
    \item Input caption: \code{"Today, we’ll explore advanced concepts in Sea Plus Plus programming, focusing on poly more fizz ‘em."}
    \item ChatGPT output caption: \code{"Today, we will delve into advanced concepts in C++ programming, specifically focusing on polymorphism and encapsulation."} 
Llama2-13B output caption: \code{"Today, we’ll delve into advanced Sea Plus Plus programming concepts, with a focus on polymorphism."}
\end{itemize}

\end{itemize}

The results demonstrate that both LLMs, ChatGPT-3.5 and Llama2-13B, effectively corrected homophones and inaccurate words in the input captions, resulting in coherent and contextually appropriate captions. 

Regarding domain-specific terminology, ChatGPT-3.5 provided a more comprehensive correction by identifying and correcting \code{"Sea Plus Plus"} to \code{"C++"}, whereas Llama2-13B retained \code{"Sea Plus Plus."}. Llama2-13B also changed \code{"focusing on"} to \code{"with a focus on."}, while this improves language use within the caption, it may not help DHH users if the video discusses bad writing styles. Additionally, if such changes by Llama2-13B change the length of the caption significantly, it will disrupt caption and video synchronization. These results indicate that while both models successfully correct general language errors, ChatGPT-3.5 shows a slightly better capability in handling domain-specific terminology. However, ChatGPT-3.5 also added \code{"encapsulation"} to \code{"polymorphism"}. While this shows better domain-specific understanding, it may not help DHH users if the video doesn't discuss encapsulation. 


\subsection{Quantitative evaluation metrics}

We employed three standard metrics to measure the performance of our caption correction pipeline: Word Error Rate, BLEU Score, and ROUGE Score. These metrics are essential for understanding how closely the generated captions match the ground truth captions.

To clarify the terms used in our evaluation, here are the definitions:

\begin{itemize}
\item Generated caption: Automatically generated caption by an assistive tool (e.g., YouTube's ASR system). 
    \item Ground truth caption: The correct caption of the video, which is generated manually. 
    \item Predicted caption: This is the caption generated by LLM (Corrected caption).
\end{itemize}

\textbf{Word Error Rate (WER)} \cite{10}: Measures the percentage of errors in the predicted caption compared to the reference caption. It accounts for incorrect, omitted, or inserted words, calculated using the formula in Equation \ref{eq:1}. A lower WER indicates a higher accuracy of the predicted captions, and a WER of 0\% indicates a perfect match.

\begin{equation} \label{eq:1}
WER = \frac{S + D + I}{N}
\end{equation}

\begin{itemize}
    \item Substitutions ($S$): Incorrect words in the predicted caption.
    \item Deletions ($D$): Missing words in the predicted caption present in the ground truth caption.
    \item Insertions ($I$): Extra words in the predicted caption, not in the ground truth caption.
    \item $N$: Total number of words in the ground truth caption.
\end{itemize}

\textbf{Bilingual Evaluation Understudy (BLEU)} \cite{11}: Calculates the precision of n-grams, n-grams are consecutive sequences of n words, measuring the percentage of n-grams in the predicted caption that matches the ground truth caption. It evaluates how well the predicted caption captures the exact word sequences of the ground truth caption. Commonly used n-grams include unigrams (single words), bigrams (two-word sequences), and trigrams (three-word sequences). The final BLEU score is calculated by aggregating the precision of unigrams, bigrams, trigrams, and four-grams, balancing the contributions of shorter and longer sequences. The BLEU score ranges from 0 to 1, with higher values indicating better quality.

\textbf{Recall-Oriented Understudy for Gisting Evaluation (ROUGE)} \cite{11}: Measures overlapping n-grams between predicted caption and ground truth caption. The ROUGE score determines the recall of n-grams, ensuring the LLM-predicted caption includes essential content from the ground truth caption. The score ranges from 0 to 1, with higher values indicating better quality. ROUGE scores are categorized as follows:

\begin{itemize}
    \item ROUGE-N: Measures overlap of n-grams.
    \item ROUGE-L: Measures the longest common subsequence (LCS).
\end{itemize}

WER measures word-level errors, which do not account for the overall meaning and context of the captions; it severely penalizes minor errors that do not change the sentence's meaning. BLEU and ROUGE help address this limitation by evaluating the precision and recall of word sequences. BLEU measures the precision of n-grams, capturing the overlap of words between the predicted and ground truth caption. However, it can struggle with longer sequences and recall. ROUGE emphasizes recall and the longest common subsequence, making it better suited for assessing the overall structure and meaning. Together, these metrics ensure that captions are not only error-free but also accurate and comprehensive, capturing both lexical and semantic similarities.

\hfill \break 
The following example illustrates how each metric evaluates different aspects of caption correction performance, as shown in Table \ref{tab:example}: 

Ground truth caption:

\code{"The quick brown fox jumps over the lazy dog."}

Predicted caption: 

\code{"The quick brown fox leaps over the lazy dog."}

\begin{table}[ht]
  \caption{Evaluation Metrics for caption correction performance}
  \label{tab:example}
  \begin{tabular}{cl>{\raggedright\arraybackslash}p{5.5cm}}
    \toprule
    Metric & Score & Explanation \\
    \midrule
    WER & 11.11\% & Measures the word-level errors, indicating an 11.11\% error rate due to the single substitution of \code{"jumps"} with \code{"leaps"}. \\
    BLEU & 0.66 & Evaluates the precision of n-grams, resulting in a lower score due to the substitution affecting the overall n-gram match. \\
    ROUGE-1 & 0.89 & Measures unigram recall, showing high similarity by capturing the overlap of individual words between the captions. \\
    ROUGE-2 & 0.75 & Measures bigram recall, reflecting the impact of the verb substitution on the sequence of two-word combinations. \\
    ROUGE-L & 0.89 & Measures the longest common subsequence, indicating strong structural similarity despite the verb change. \\
    \bottomrule
  \end{tabular}
\end{table}

\section{Results}

Leveraging LLMs to improve the quality of ASR video captioning tools has shown to be a promising approach. In our evaluation, we tested two different LLMs, ChatGPT-3.5 and Llama2-13B, using our dataset of 52 videos. The experiments were conducted using Google Colab, specifically utilizing its Python notebook environment. The models' performances were assessed using standard natural language processing metrics: WER, BLEU, and ROUGE scores, as detailed in Table \ref{tab:results}. 

Our results demonstrate that ChatGPT-3.5 (WER: 9.75\%) shows a significant improvement over the original Youtube-ASR-Caption system (WER: 23.07\%), highlighting its superior accuracy in reducing errors per word. Additionally, the BLEU score for ChatGPT-3.5 is notably high at 0.85 compared to the original Youtube-ASR-Caption system BLEU score of 0.67, suggesting superior precision in n-gram matching and, thus, higher accuracy in caption generation. ChatGPT-3.5 achieved a high ROUGE-1 score of 0.98, similar to the Youtube-ASR-Caption system, indicating a strong recall of unigrams (single words). Moreover, ChatGPT-3.5 slightly outperforms the ROUGE-2 (0.97) Youtube-ASR-Caption system.

Llama2-13B, while performing well in certain aspects, did not achieve the same overall performance level as ChatGPT-3.5. Llama2-13B did not improve the WER compared to the Youtube-ASR-Caption system. This increase in WER can be attributed to Llama2-13B's tendency to improve the language of the text, which results in better readability and coherence but does not match the original captions exactly. For example, Llama2-13B might change "focusing on" in the original caption to "with a focus on" as shown in section \ref{captionChallenge},  which improves the language but increases the WER because it deviates from the ground truth wording.  The relatively lower BLEU score (0.68) and ROUGE scores (ROUGE-1: 0.90, ROUGE-2: 0.82, ROUGE-L: 0.88) for Llama2-13B indicate that, while it captures some aspects of the captions correctly, it struggles with precision and recall compared to ChatGPT-3.5.

\begin{table*}[ht]
  \caption{Results – Comparison of Caption Correction Performance: YouTube's ASR System vs. ChatGPT-3.5 and Llama2-13B}
  \label{tab:results}
  \centering
  \begin{tabular}{cccccl}
    \toprule
    \makecell{Metric} & \makecell{WER} & \makecell{BLEU} & \makecell{ROUGE-1} & \makecell{ROUGE-2} & \makecell{ROUGE-L} \\
    \midrule
    \makecell{Youtube-ASR-Caption\\ } & 23.07 & 0.67 & 0.98 & 0.96 & 0.98 \\ \\
    Llama2-13B & 24.42 & 0.68 & 0.90 & 0.82 & 0.88 \\
    \textbf{ChatGPT-3.5} & \textbf{9.75} & \textbf{0.85} & \textbf{0.98} & \textbf{0.97} & \textbf{0.98} \\
    \bottomrule
  \end{tabular}
\end{table*}

\section{Conclusion}
In conclusion, integrating LLMs into caption correction systems can significantly enhance the accuracy and coherence of captions, improving accessibility for the DHH community. Our evaluation shows that ChatGPT-3.5 outperforms Llama2-13B regarding caption quality, capturing both individual words and longer sequences effectively and ensuring the generated captions are detailed and contextually accurate.

However, a limitation of this project is that LLMs might miss or misinterpret voice intonations, cultural references, and idioms that manual captioning can understand and convey accurately. For instance, a human captioner can appropriately interpret the emphasis in \code{"Really?!"} to express surprise, while an LLM might transcribe it as \code{"Really"}. Similarly, cultural references like \code{"Diwali"} (the Hindu festival of lights) or phrases like \code{"namaste"} (a traditional Hindu greeting) could be misunderstood by LLMs but accurately captured by human captioners. To address these limitations, we propose employing multi-modal LLMs and incorporating ML models designed to understand complex human communication nuances. 
\section{Future Work}
Future research and development efforts can enhance the effectiveness of video captioning through several key areas. Expanding our current dataset beyond YouTube ASR system-generated captions to include other widely used platforms like Microsoft Teams and Zoom would increase the applicability of our solutions to the DHH community.

In addition, considering the potential scalability of our approach is crucial for maximizing its impact. Investigating how well the LLM-based captioning system can handle larger datasets and diverse input types will be essential for scaling up the solution to different environments and applications. Exploring the deployment of this solution in low-resource settings or environments with limited computational capabilities is a necessary step to ensure accessibility across a broad range of users. Optimization of model efficiency, such as using quantization techniques or lighter model variants, could further facilitate the scalability of this approach to a production-level tool.

The applicability to other assistive technologies is another avenue worth exploring. Beyond video captioning, the methods developed here could extend to applications like real-time speech-to-text systems used in classrooms or workplaces, benefiting not only the DHH community but also those with auditory processing disorders or language learners. Integrating LLMs for enhanced subtitle generation in augmented reality (AR) or virtual reality (VR) environments could also pave the way for immersive and accessible experiences, improving real-time communication. Exploring these broader applications will further demonstrate the flexibility and impact of LLM-based solutions in assistive technology.

\begin{figure}
\begin{center}
    \includegraphics[alt = {Preliminary experiments to evaluate LLMs' performance on code-switching challenges in ASR video captions demonstrate that Llama-2-13B and ChatGPT-3.5 successfully corrected the caption, showcasing their ability to handle mixed-language inputs and produce accurate English captions.}, width=8.5cm]{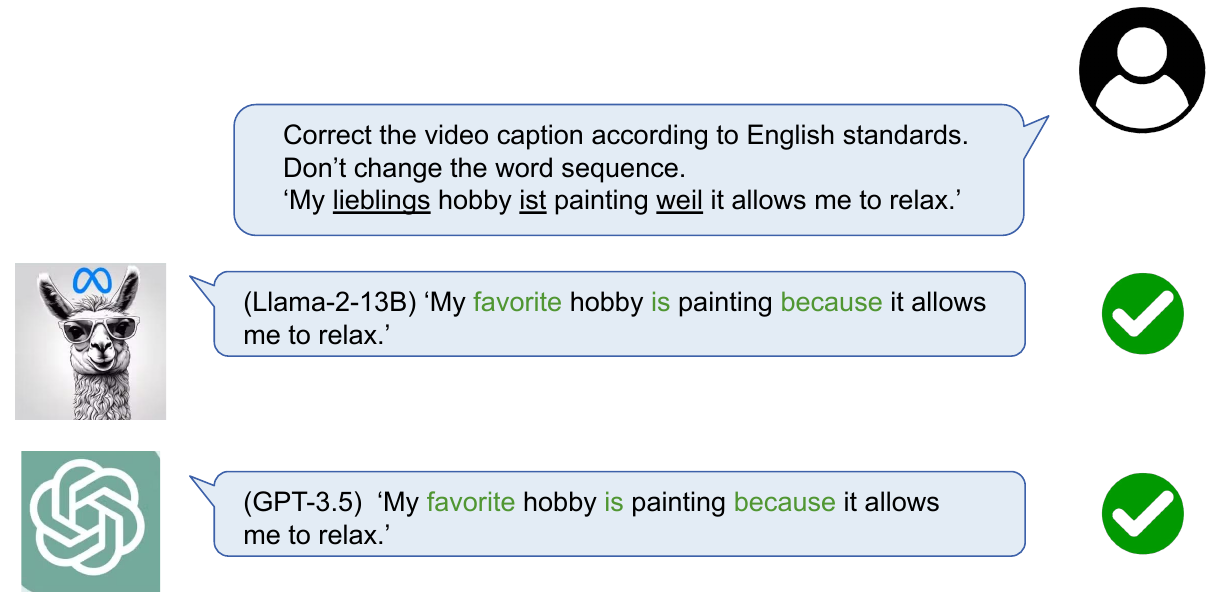}
  \caption{Preliminary experiments to evaluate LLMs' performance on code-switching challenges in ASR video captions demonstrate that Llama-2-13B and ChatGPT-3.5 successfully corrected the caption, showcasing their ability to handle mixed-language inputs and produce accurate English captions.}%
    \label{fig:codeSwitching}%
\end{center}
\end{figure}

Addressing code-switching is another essential area of focus. YouTube’s ASR system feature currently handles only one language at a time, which presents a significant limitation. Existing code-switching ASR datasets such as \cite{20}, \cite{21}, \cite{22} handle switching between two languages only, which is not suitable for evaluating LLM performance on audio containing more than two languages. Curating a dataset to address this challenge requires using tools such as Whisper by OpenAI \cite{19}, designed to handle multiple languages within a single audio stream. We conducted some preliminary experiments to evaluate LLMs' performance for the code-switching challenge, a sample of the experiments is shown in figure \ref{fig:codeSwitching}. Our results show that ChatGPT-3.5 and Llama2-13B can successfully handle mixed-language inputs and produce accurate English captions. 

Furthermore, experimenting with more advanced prompt engineering techniques, such as context-aware prompting. Context-aware prompting is a prompt engineering technique that includes relevant information in the prompt to improve LLM responses. Contextual information includes but is not limited to the domain of the video and its metadata. 

Lastly, user experiments on the pipeline will be conducted with the DHH community to gather feedback and make necessary modifications. This feedback can also improve LLM performance through reinforcement learning, ensuring a user-centered solution that effectively meets their needs. 

\section*{Acknowledgements}
This work was supported by the Integrated AI in Teaching at the University of Stuttgart project (IKILeUS), funded by the German Federal Ministry of Education and Research (BMBF).


\end{document}